# Plant Disease Detection Using Deep learning

**Anjaneya Teja Kalvakolanu**
Department of Computer Science, California Polytechnic State University

*Abstract-* Plant disease detection is a huge problem and often require professional help to detect the disease. This research focuses on creating a deep learning model that detects the type of disease that affected the plant from the images of the leaves of the plants. The deep learning is done with the help of Convolutional Neural Network by performing transfer learning. The model is created using transfer learning and is experimented with both resnet 34 and resnet 50 to demonstrate that discriminative learning gives better results. This method achieved state of art results for the dataset used. The main goal is to lower the professional help to detect the plant diseases and make this model accessible to as many people as possible.

*Index Terms*- Deep Learning, Plant disease detection, Discriminative Learning, Resnet 34, Resnet 50, Plant village.

## I. INTRODUCTION

This Research is focused on Detection of diseases that plant is affected by using deep learning. The images of affected plant leaves are taken as input and are sent through a network trained using transfer learning from resent 34 and resent 50. We considered three plants for this research Potato, Tomato, Bell pepper. In particular the diseases with plants considered are early blight, mosaic virus, late blight, leaf mold, bacterial Spot, Leaf curl, Target spot ,spider mite. In addition we also considered Healthy plants as a class. The classes are divided based on plant and disease type since each plant has a different leaves shape and many diseases appear similar on same plant. We used transfer learning with both resnet 34 and resnet 50 as our base model. We trained the model with around 4000 images with discriminative learning . we achieved an accuracy of 99.44% which is current state of art for this dataset.

## II. BACKGROUND

There has been fast paced research for Plant disease detection using images in the recent years. Penn state has created their own Plant village has developed their own dataset to create a deep learning model for detection of diseases in plants. There have been many models developed and tested on the plant village dataset. Methods like segmentation are used to detect the affected part of the leaves and they are classified using the deep learning methods. Vgd is used to classify the images and they achieved an accuracy of 97.2. In order to develop image classifiers for the purposes of plant disease detection , there is a requirement of large and verified dataset of healthy and diseased plants. Until plant village such dataset is not available.

Implementation of RGB and Gray scale images in plant disease detection in leaves a study by Padmavathi and Thangadurai have given the comparative results of RGB and Gray scale images in disease finding process from leaves. Color is an important feature in detecting the infected leaves. Color also identifies the intensity They have considered Grayscale and RGB images and used median filter for image enhancement and segmentation for extraction of the diseased portion which are used to identify the disease intensity. They identified 13 different diseases from the healthy leaves with the capability to differentiate from the surroundings from the leaves.

## III. EXPERIMENT

The transfer learning is done using both resent 34 and resnet 50. The data set obtained has over 4000 images that are divide in to classes based on the plant type and disease type. The pre processing include data augmentation. The images are flipped, warped, transformed to make the dataset more generic.

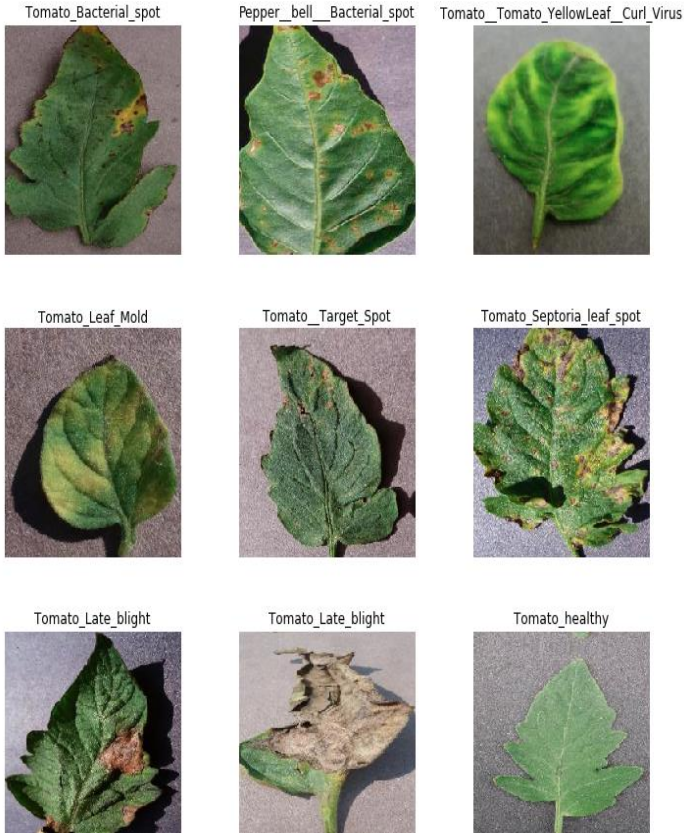

The dataset is turned in to a batch of 64 images, each image is of size 224 * 224. The normalization is performed with the image net stats since the imported resnet model is trained on the image net .The loss function used is categorical cross entropy.

## IV. DISCRIMINATIVE LEARNING

The CNN model created from the resnet is trained by using the discriminative learning. The initial training is done only for the newly added layers after performing the hyper parameter tuning for learning rate.

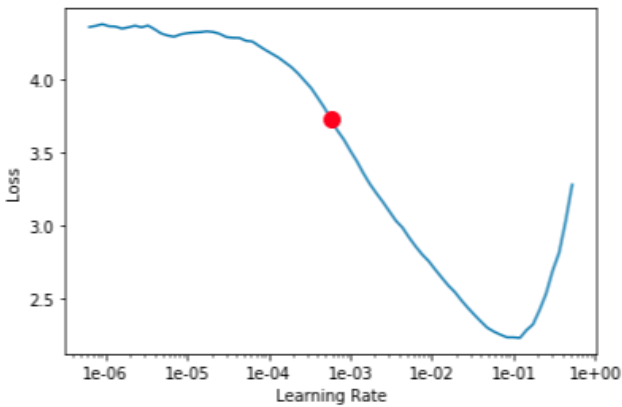

The model is trained with 3e-03 learning rate only for the newly added layers.

| epoch | train_loss | valid_loss | accuracy | time |
|---|---|---|---|---|
| 0 | 1.129517 | 0.374790 | 0.877635 | 2:41:30 |
| 1 | 0.275372 | 0.119957 | 0.960989 | 2:38:58 |
| 2 | 0.145123 | 0.059501 | 0.982796 | 2:39:36 |
| 3 | 0.095666 | 0.041671 | 0.987158 | 2:40:10 |

This initial training worked well and we achieved an accuracy of 98.7 percent after training for 4 epochs with a batch size of 64. Once the training is done to perform discriminative learning all the other layers are unfrozen and the training is performed with higher learning rate for the initial layers and the lower learning rate for he end layers. The end layers are already good at detecting the features of the images sent in so it makes sense to train them with low learning rate. The initial layers have only the pre trained weights so they should be trained to be able to detect the features of the images so high learning rate is used. This technique worked in improving the accuracy further more. Hyper parameter tuning is performed again.

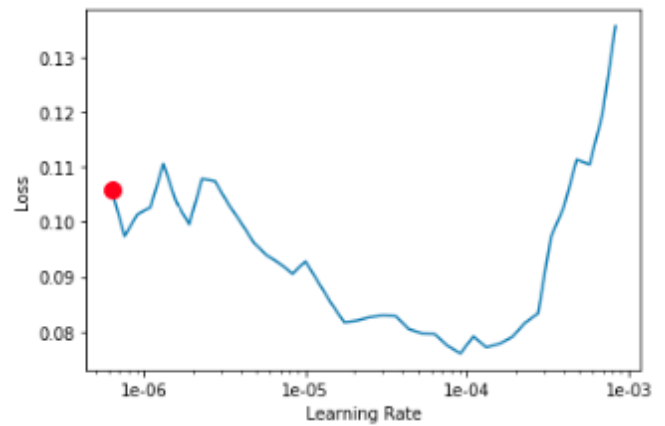

The next training is done with learning rate between 1e-05 and 1e-04 with high learning rate for initial layers and then distributed the learning rate with the multiples for the next layers between start and end learning rate. This method gave us an accuracy of 94.4 percent which is the current state of art.

| epoch | train_loss | valid_loss | accuracy | time |
|---|---|---|---|---|
| 0 | 0.032650 | 0.023584 | 0.994185 | 3:22:56 |
| 1 | 0.042813 | 0.022023 | 0.993215 | 3:23:48 |
| 2 | 0.032231 | 0.019895 | 0.994427 | 3:24:00 |

## V. RESULTS

We tried working with both resnet 34 and resnet 50 even though resnet 50 achieved better accuracy than resnet 34 (99.1%)

took very less time to enter achieve 99 accuracy. Since the validation loss and training loss both are lowering and the accuracy is not falling plotted the training and validation losses which makes sure that the model is not over fitted. The top most losses are plotted in the format actual/predicted/loss/probability and it is clear that most of them happened due to errors in dataset and even human can go wrong with them. The confusion matrix is also plotted.

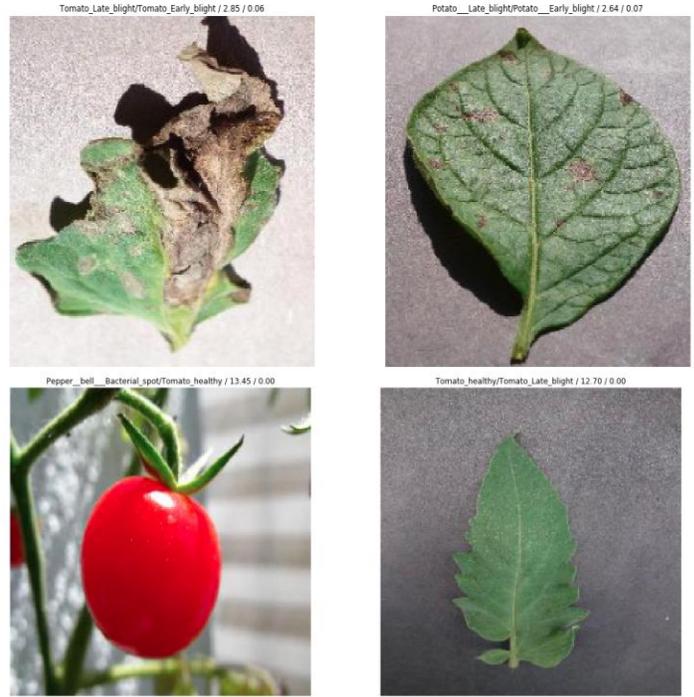

## VI. CONCLUSION

Our proposed model with Discriminative learning rate achieved an accuracy of 99.44 percent which is the current state of art for this dataset. New data can be added to train the model and this will not lower the accuracy because of the regularization and deeper layers of resnet 50. This work can be further extended by adding more data and training the network.

Top most losses : Predicted/Actual/Loss/Probability

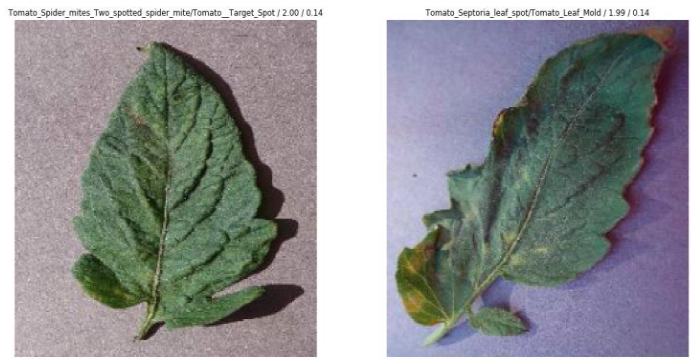